\documentclass[conference]{IEEEtran}
\IEEEoverridecommandlockouts
\usepackage{array}
\usepackage{cite}
\usepackage{float}
\usepackage{amsmath,amssymb,amsfonts}
\usepackage{placeins}
\usepackage{graphicx}
\usepackage{epstopdf}
\usepackage{textcomp}
\usepackage{xcolor}
\usepackage{url}
\usepackage{longtable}
\usepackage{bbm,bm} %
\usepackage{subcaption}
\usepackage{amsthm}
\usepackage{booktabs}
\usepackage{multirow}
\usepackage{tabularx}
\usepackage{algorithm}  
\usepackage{algorithmicx}
\usepackage{enumitem}
\usepackage{algpseudocode}

\def\BibTeX{{\rm B\kern-.05em{\sc i\kern-.025em b}\kern-.08em
    T\kern-.1667em\lower.7ex\hbox{E}\kern-.125emX}}

\usepackage{xspace} 

\newcommand{\eg}{\emph{e.g.,}\xspace}

\newcommand{\see}{\emph{see}\xspace}

\begin{document}

\title{Deep Reinforced Attention Regression for Partial Sketch Based Image Retrieval

}

\author{
\IEEEauthorblockN{Dingrong Wang, Hitesh Sapkota, Xumin Liu, Qi Yu}
\IEEEauthorblockA{Rochester Institute of Technology}
\IEEEauthorblockA{\{dw7445, hxs1943, xumin.liu, qi.yu\}@rit.edu}
}

\maketitle
\begin{abstract}

Fine-Grained Sketch-Based Image Retrieval (FG-SBIR) aims at finding a specific image from a large gallery given a query sketch. Despite the widespread applicability of FG-SBIR in many critical domains (\eg crime activity tracking), existing approaches still suffer from a low accuracy while being sensitive to external noises such as unnecessary strokes in the sketch. The retrieval performance will further deteriorate under a more practical on-the-fly setting, where only a partially complete sketch with only a few (noisy) strokes are available to retrieve corresponding images. 
We propose a novel framework that leverages a uniquely designed deep reinforcement learning model that performs a dual-level exploration to deal with partial sketch training and attention region selection.
By enforcing the model's attention on the important regions of the original sketches, it remains robust to unnecessary stroke noises and improve the retrieval accuracy by a large margin. To sufficiently explore partial sketches and locate the important regions to attend, the model performs bootstrapped policy gradient for global exploration while adjusting a standard deviation term that governs a locator network for local exploration. The training process is guided by a hybrid loss that integrates a reinforcement loss and a supervised loss. A dynamic ranking reward is developed to fit the on-the-fly image retrieval process using partial sketches. The extensive experimentation performed on three public datasets shows that our proposed approach achieves the state-of-the-art performance on partial sketch based image retrieval. 

\end{abstract}


\section{Introduction}
Due to the large amounts of touch-screen devices and their broad usage, sketch-related computer vision applications have drawn increasing attention nowadays. Consequently, a series of interesting research problems have emerged, including 
sketch recognition, sketch representation/interpretation, sketch generation, and sketch-based image retrieval (SBIR) \cite{Yu2016,Song2017,Pang2019,Dey2019,Dutta2019,Collomosse2019}. Among them, SBIR, especially fine-grained SBIR, is of particular interest \cite{Dey2019,Dutta2019,Collomosse2019} because of its important applications in both commercial and public safety domains. For commercial usage, a customer can buy items online by simply drawing sketches of items using a smartphone screen. For the public safety applications, it provides additional support to track criminals by drawing their sketches based on the description and searching them in the police criminal database. 

Category level SBIR is well studied in existing works~\cite{Yu2016,Dey2019,Dutta2019}. In this technique, based on a query sketch, it retrieves images of the same category as the query. However, the domain gap between sketches and images remains as a key technical challenge. This is because a sketch captures only object shape/contour information, which does not contain other relevant fine-grained details in an image, such as color and texture \cite{Song2017}. There are three common ways to deal with this problem. First is sketch-image generation combining with image recognition\cite{Guo2017}. But this method's performance depends heavily on the performance of sketch-image generation, where uncertainty is extremely high. The second approach is sketch-image hashing\cite{Xu2018}, where sketch and image are encoded by a deep neural network into hashing codes, then by calculating the Hamming distance between query sketch and each category's image hashing cluster, we can decide which cluster's images should be retrieved based on the query sketch. The last approach is finding a common sketch-image  embedding layer\cite{Dey2019}, which aims to find a common hidden space to align both sketch and image embedding vectors. Although these methods can reach high classification accuracy, category level SBIR is usually not suitable in many real  applications, where people expect to see a more precise retrieval result instead of a category of images.  FG-SBIR provides a promising direction to fulfill this demand. 

\begin{figure*}[t!]
\centering
\vspace{-16mm}
 \includegraphics[width=\textwidth]{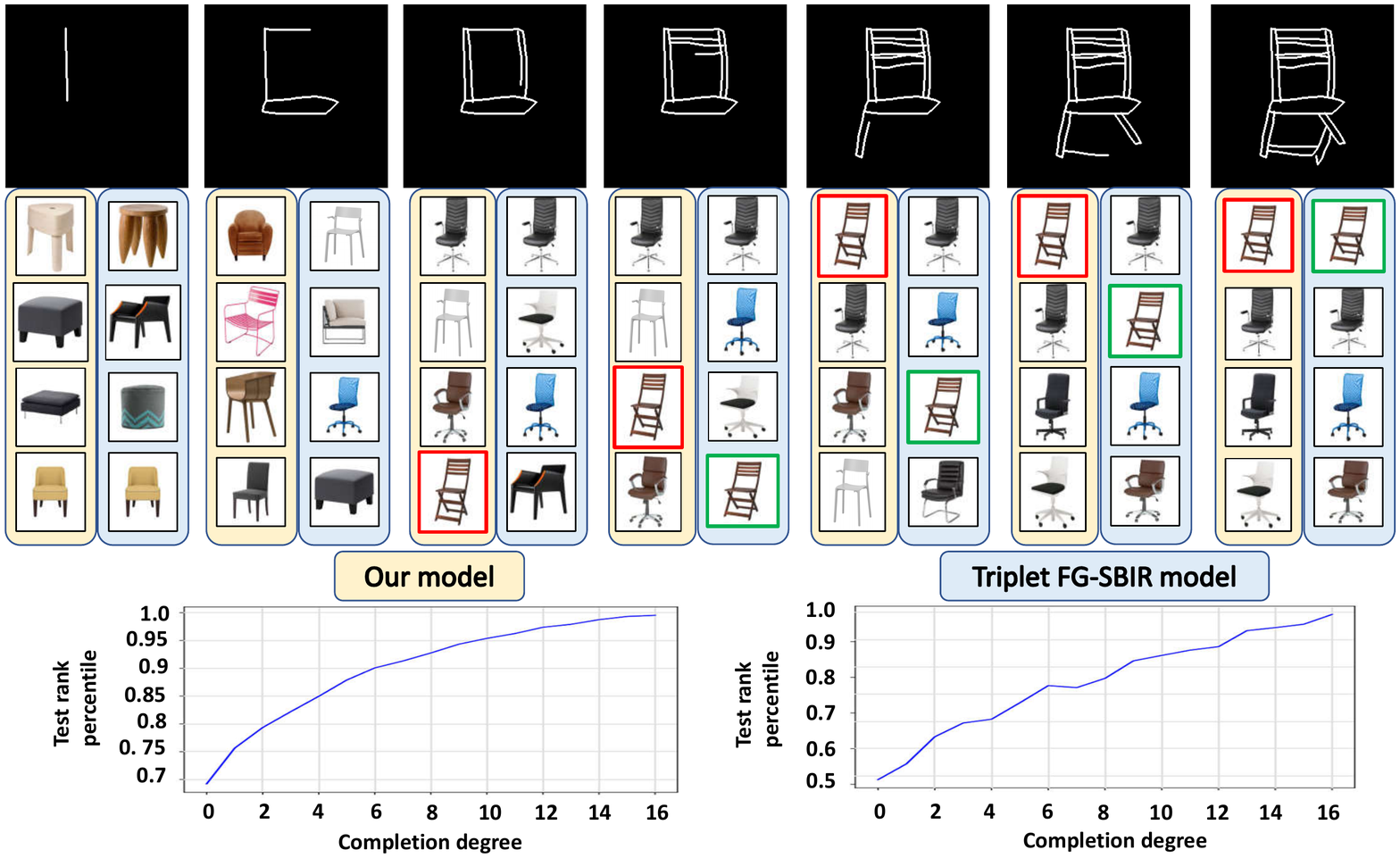}
\vspace{-16mm}
\caption{Partial Sketch Retrieval Performance Comparison Between DARP-SBIR and a Triplet Network Model\label{fig:partialretrieval}. The top row shows a sequence of partially complete sketches with increasing degree of completeness; the second row shows the corresponding retrieved list of images by DARP-SBIR (yellow) and Triplet network (blue); the last rows shows the ranking percentile of the matching image (\eg 0.9 means the matching image is ranked in the top 10\% among all candidate images).}
\label{system}
\vspace{-2mm}
\end{figure*}

Compared to the category level SBIR, 
FG-SBIR can explore more fine-grained details from both sketch and image sides and get a more precise retrieval result. FG-SBIR is initially addressed in \cite{Li2014}, which employed a deformable part-based model (DPM) representation and graph matching. However, the model performs poorly (in terms of retrieval accuracy) whenever the situation gets complicated (\eg adding more abstract sketches or images from unknown categories). To deal with more complicated settings, deep learning has been leveraged to tackle FG-SBIR, which aims to learn both feature representation and a cross-domain matching function jointly \cite{Song2017,Pang2019,Lei2019}. In this research area, there are broadly two lines of work: Siamese CNN network with a verification loss \cite{Sangkloy2016} and triplet CNN with a triplet ranking loss~\cite{Pang2017}\cite{Song2017}\cite{Yu2016}. The difference in those works lies in the network structure and the image branch. A Siamese network takes extracted edge maps as input in the image branch for alignment with the sketch branch, while a triplet network uses the original image as input, which contains more information. The problem with these models is that to reach a good retrieval performance, a complete sketch is needed. 
In practice, like an online shopping app, people like to retrieve what they want in just a few strokes, rather than the whole detailed sketch. However, these models can only predict on the full sketch and have poor performance when generalizing to partial sketch cases. Simply adding partial sketches into the training set is not applicable, as this jeopardizes the overall training performance.

Existing work to provide image retrieval based on partial sketches is fairly scarce. One exception is the model that combines a triplet network with a reinforcement learning (RL) agent to predict the embedding of a partial sketch on the fly~\cite{Bhunia2020}. 
In this model, a partial sketch is regarded as a state of a particular time step from a full episode. The reward signal here does not correspond to the final optimization objective adopted by supervised training. Instead, it belongs to the total cumulative reward in a whole episode, which is more natural for the on-the-fly setting. As a result, RL training for partial sketches has the potential to get better results than the supervised training. While the model shows a promising trend to support partial sketch based retrieval, the performance is still much lower than the full sketch based result. The result gets even worse when the completion degree of a partial sketch is low as the model becomes more sensitive to the noisy stokes. 

To address the fundamental challenges as outlined above, we propose a novel framework that employs Deep reinforced Attention Regression for Partial Sketch Based Image Retrieval (DARP-SBIR). The proposed framework aims to deal with partial sketches while being robust to noisy stokes to further advance the state of the art in FG-SBIR and support the practical on-the-fly image retrieval. While a RL agent adopted by existing works can capture the temporal dynamics among different stokes in a sketch to predict the embedding of a final sketch~\cite{Bhunia2020}, the incomplete sketches containing noisy strokes along with a potentially large image repository make the reward signal very sparse and delayed. The proposed attention mechanism aims to identify the most important regions in the sketch while ignoring the unnecessary noisy strokes. To tackle the sparse and delayed reward signals, the framework employs a novel bootstrapped policy gradient with dual-level exploration that performs a global deep exploration with a multi-head locator network and local exploration by dynamically adjusting the covariance of the Gaussian policy. The state space of the RL agent is maintained by a RNN network that aggregates both the attended location and the currently available sketch, 
The hidden state from the RNN can also be used to predict the sketch embedding to formulate a supervised learning loss, which is used to training the RNN (and other component networks in the framework) to more accurately capture the hidden state space. Finally, a dynamic ranking reward function is developed to allow the agent to pick up the relatively weak reward signal in the early phase and then focus on the higher reward once it is better trained.

Figure~\ref{fig:partialretrieval} compares the partial sketch retrieval performance between the proposed DARP-SBIR framework and a classical triplet network based model \cite{Yu2016}. From the second row, we can see that with just a few stokes (third column), our model can already rank the matching image in its top-4 list. DARP-SBIR also puts the matching image on the very top of its retrieval list multiple steps ahead of the Triplet network, which requires the complete sketch to do so.  It is worth to note that only a subset of stokes are included from a whole rendering episode so the actual gap between these two models is actually even bigger. The bottom row quantitatively evaluate the partial sketch retrieval performance using the ranking percentile vs. completion degree of sketches (defined in Section~\ref{sec:exp}). Our main contributions are summarised as follows:
\begin{itemize}[noitemsep,topsep=0pt,leftmargin=*]
    \item  We design a novel framework that employs deep reinforced attention regression to support partial sketch based on-the-fly image retrieval. 
    \item To handle sparse and delayed reward signals due to incomplete sketches and noisy stokes, the RL agent performs dual-level exploration that combines deep global exploration through bootstrapped policy gradient and local exploration through dynamically adjusted Gaussian covariance. 
    \item The state space of the attention RL agent is maintained through a RNN that dynamically aggregates both the current attended region provided by the agent and the current available sketch. 
    \item A hybrid loss function is developed that combines a dynamic ranking reward for training the RL agent and a supervised learning loss for updating the RNN.
\end{itemize}
 Extensive experiments have been conducted on three public sketch datasets to show that our proposed DARP-SBIR framework achieves the state-of-the-art performance on both complete and partial sketch based image retrieval.

\section{Related Work}
\subsection{Category Level SBIR}
Category-level sketch-image retrieval has been extensively studied by existing works \cite{Bui2018,Cao2011,Collomosse2017,Dey2019}. There are three mainstream methodologies in the field of category-level SBIR: (1) sketch-image generation and image recognition, (2) sketch-image hashing, and (3) sketch-image common embedding. For the first group of methods, representative works like \cite{David2017} use a sketch-RNN to perform sketch-image generation and then use image recognition to find the matching images, which can be unreliable because the sketch-RNN may provide poor multi-class sketch-image generation. In the second group of work, Liu et al \cite{Liu2017} apply a CNN-RNN two branch model to deal with images and sketches, respectively. In their approach, a CNN extracts static information from images, while a RNN extracts temporal information from strokes. Finally, both sketches and images can be encoded into hashing vectors and SBIR is achieved by comparing the hamming distance between the query sketch and different image clusters. The most representative work in the third group mainly leverages a triplet network, where a triplet input consisting of one sketch, one positive image, and one negative image is fed into the network to transform into hidden vectors in the common embedding layer. Then, through a triplet loss, the sketch embedding vector should be close to the positive image embedding vector while moving away from the negative image's embedding vector, just like clustering in the hidden space \cite{Pang2019, Dey2019}. In this way, by performing a simple $l_2$ distance search in the hidden space with the query sketch's embedding vector, we can pick the closest image embedding as the matching image. Since the target is a set of images instead of a single image, category level SBIR may not be suitable for certain applications because its retrieval result is not precise. However, it provides a good starting point for FG-SBIR.
\subsection{Fine-grained SBIR}
Fine-grained SBIR is a more recent research area, which is less investigated than category level SBIR in the sketch analysis field, partly due to the challenge in data collection. Initially, it was addressed by a deformable part-based model representation and graph matching~\cite{Li2014}. After that, a number of deep learning approaches have been applied to find the embedding space for both sketches and images \cite{Yu2016,Song2017,Pang2019,Pang2017}. Two branches training with edge map and verification loss has been replaced by a triplet network with a ranking loss \cite{Song2017,Pang2019}. Nowadays, FG-SBIR is researched using more complicated network structures and carefully designed losses. For example, Yu et al.\cite{Yu2016} proposed a deep triplet-ranking model for instance-level FG-SBIR. Then, this paradigm is replaced by hybrid generative-discriminative cross-domain image generation~\cite{Pang2017}, combined with a more sophisticated triplet loss. Song et al. \cite{Song2017} propose a model, which is also a triplet CNN. But with the introduced multi-scale coarse-fine semantic fusion and HOLEF loss, their model is much more effective than previous ones.  Although existing models can reach better accuracy for complete sketches, they cannot readily generalize to incomplete ones, which is the focus of our model. 

\subsection{Reinforcement Learning based SBIR}
While there has been much research effort in leveraging reinforcement learning
(RL) \cite{Kaelbling1996} in various computer vision problems, RL is not commonly used for SBIR with few exceptions. Ayan et al. propose a partial sketch training procedure, where RL is leveraged to trade-off between sketch recognisability and the number of strokes to observe~\cite{Bhunia2020}. It introduces a discount weight and negative rewards for the number of strokes. Although the model is able to handle partial sketches, the retrieval performance is still lower comparing to the full sketch FG-SBIR models mentioned before. Further, the model itself is not robust to disturbing strokes. In our model, we use a novel attention mechanism combined with carefully designed dual-level exploration to improve the retrieval accuracy and avoid disturbing strokes by selecting important regions to feed into the network.

Effective deep exploration is crucial in RL tasks where the environment has a large action and state spaces. Various attempts have been made for effective deep exploration. Inspired from the idea of Thompson sampling \cite{Thompson1933}, various posterior sampling based approaches have been proposed \cite{Guez2012, Osband2013, Strens00abayesian} to facilitate deep exploration. However, these techniques face the problem of inefficiency and intractability. To overcome the problems in previous approaches, randomized value function based approaches have been proposed. Based on this idea, both linearly \cite{Osband2016v1} as well as non-linearly \cite{Osband2016} parameterized value functions are proposed. Relying on a non-linearly parameterized value function, Osband et al. \cite{Osband2016} developed Bootstrapped DQN, which combines DQN and the bootstrapping technology. It performs well in multiple Atari games and boosts the performance significantly. 
However, the training time and computing resource cost pose a key bottleneck in a continuous action space, like attention regression in SBIR. To solve this problem while still leveraging the benefit of bootstrapping, we propose bootstrapped policy gradient training to alleviate the computation cost and further boost the performance through effective local exploration.

\begin{table}[htbp]
\caption{Notations with Descriptions}
\vspace{-2mm}
\small
\begin{center}
\begin{tabular}{|c|p{6.3cm}|}
\hline
{\bf Notation} & {\bf Description}  \\ [0.5ex] 
\hline \hline
${\bf x}_i$& the $i$-th input partial sketch  \\
\hline
${\bf e}_i$& matching image embedding corresponding to ${\bf x}_i$\\
\hline
${\bf g}_{i,t}$& glimpse vector at time step $t$ for ${\bf x}_i$\\
\hline
${\bf a}_{i,t}$& embedding vector of ${\bf x}_i$ at time step $t$\\
\hline
${\bf v}_{i,t}$& glimpse location at time step $t$ for ${\bf x}_i$\\
\hline
${\bf h}_{i,t}$& hidden state vector at time step $t$ for ${\bf x}_i$\\
\hline
$\boldsymbol{\omega}_{i,t}^k$& binary masks for ${\bf x}_i$ generated from locator head $k$ at time step $t$\\
\hline
$R_{i,t}$& reward signal for ${\bf x}_i$ at time step $t$\\
\hline
$\eta_{t}$& threshold to adjust the dynamic ranking award\\
\hline
$\theta_g, \theta_h, \theta_a, \theta_v$& parameters of the glimpse, RNN, action, and locator networks\\
\hline
$M$& the number of total episodes\\
\hline
$N$& the number of partial sketches in one episode\\
\hline
$T$& the number of total time steps per episode\\
\hline
$K$& the number of locator heads\\
\hline
\end{tabular}
\label{tab: symbol_table}
\end{center}
\vspace{-4mm}
\end{table}

\section{Methodology}
We aim to develop a partial sketch based image retrieval framework that is capable of overcoming the well-known limitations for FG-SBIR: low retrieval accuracy and vulnerability to disturbing strokes. To fulfill these purposes and make our framework easy to adjust to partial sketches, we design a novel bootstrapped policy network, which enforces dynamic attention on the original sketch directly. Specifically, we divide the model training into two phases. In the first phase, we train the powerful backbone triplet Inception V3 model coupled with a triplet loss to get the target image embedding. In the second phase, we use the partial sketch based retrieval framework, which consists of a glimpse network, an action network, and a locator network to enforce dynamic visual attention on input sketches directly by selecting important attention regions. The selected attention location is processed by the glimpse network along with the current (partial) sketch, which will be sent to a RNN network to generate a hidden state vector. The state vector can be directly used by the action network to generate the sketch embedding. Meanwhile, the state is also used by the locator network to produce the updated attention location for the next time step. This dynamic attention mechanism is the core of our design, which could avoid noisy strokes from input sketches, as well as reduce the input dimensionality for more efficient training. Previously generated target image embedding is used to calculate the ranking reward and the reinforcement loss for bootstrapped policy gradient training to teach the model to look into the important regions. We further leverage a supervised loss to collectively train the glimpse, RNN, and action network to generate the sketch embedding as close to the ground truth image embedding as possible. Table \ref{tab: symbol_table} summarizes the major notations used throughout the paper. 



\begin{figure}[h!]
\vspace{-2mm}
  \centering
  \includegraphics[width=0.45\textwidth]{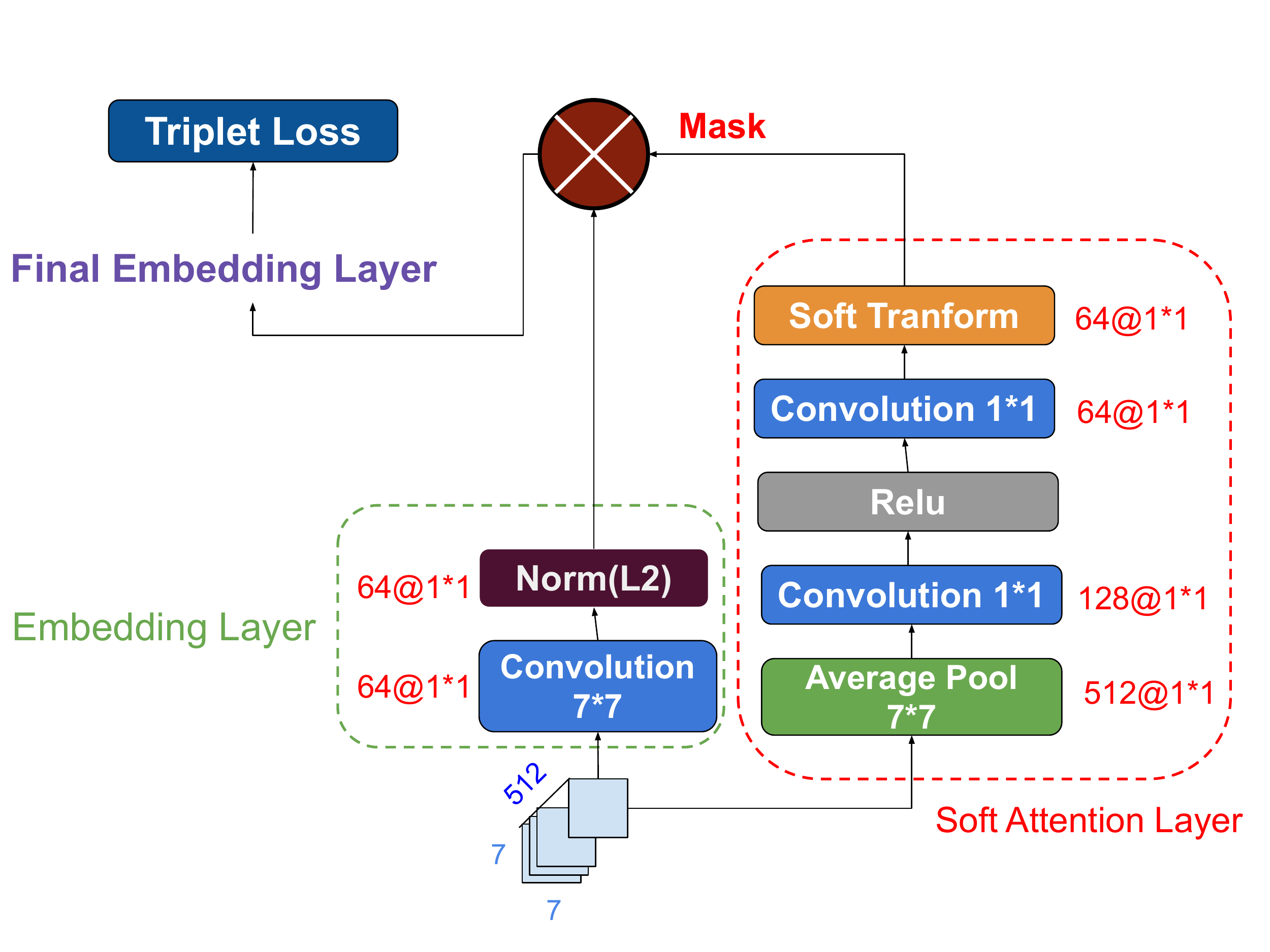}
\vspace{-2mm}
  \caption{Structure of the Embedding Network}
  \label{triplet}
\vspace{-2mm}
\end{figure}%

\begin{figure*}[t!]
  \centering
  \includegraphics[width=0.8\textwidth]{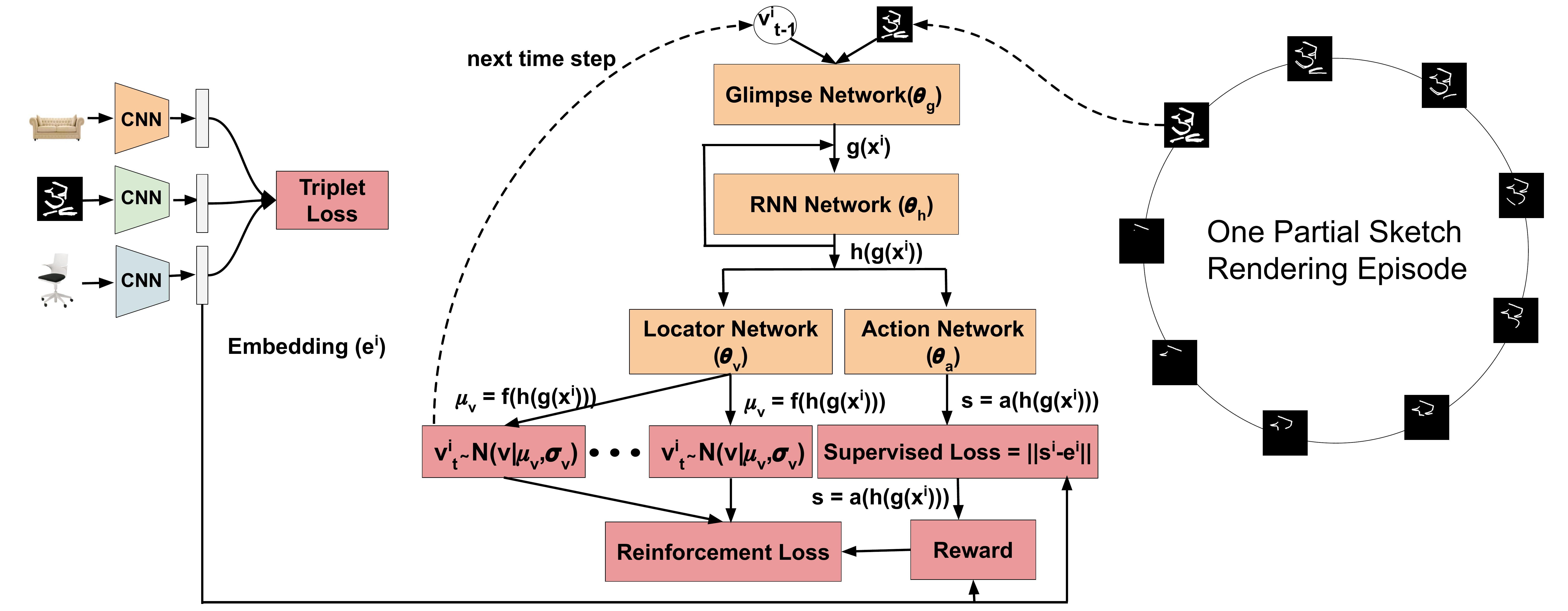}
  \caption{Overall Architecture of DARP-SBIR\label{fig:darp-arch}}
\vspace{-2mm}
\label{Framework}
\end{figure*}

\subsection{Image Embedding Network}

In order to get pre-trained ground-truth image embedding vectors, we first train a triplet network consisting of three branches with a shared set of parameters. The input is passed in the form of anchor, positive and negative samples. The goal is to minimize the distance between the anchor and the positive samples while
maximizing its distance from the negative sample. In our case, the
anchor is a sketch, the positive sample is an image corresponding to that sketch, called the ground truth image, while the negative sample is a random image not corresponding to that sketch. For the network structure, as Figure \ref{triplet} shows, we add a fully connected embedding layer with 64 neurons on the top of the feature extractor (which is the backbone model of Resnet or InceptionV3) to get the embedding vector. We then normalize the output embedding vector in $l_2$ form. On the other side, we use a soft attention layer to extract effective part for that embedding vector to form a final embedding vector to be utilized by the main framework. For network training, we apply a triplet loss, which is formulated as follows. Define a triplet $\{a,p,n\}$, where $a$ is anchor, $p$ is a positive example, $n$ is a negative example. Assume that $a$ and $p$ are from the same category, while $n$ is not. Let $\psi(\cdot)$ and $\phi(\cdot)$ denote the embedding vectors for a sketch and image, respectively. Then, $l_2$ distance is used to calculate their difference, where $\delta_+$ and $\delta_-$ denote the distances to the positive and negative instances, respectively. 
Finally, the triplet loss is defined as
\begin{equation}
    \mathcal{L}_{tri} = \frac{1}{N}\sum_{n=1}^N \max\{0,\mu+\delta_+^n-\delta_-^n\}
\end{equation}
where $\mu$ is a hyper-parameter that defines a margin.



\subsection{Deep Reinforced Attention Regression Network}
In this section, we present the architecture (\see Figure~\ref{fig:darp-arch}) of the proposed deep reinforced attention regression network for partial sketch based image retrieval (DARP-SBIR), which includes four key components: a glimpse network, a RNN network, an action network, and a locator network. We describe the first three components in this section and present the locator network next section that focuses on its unique dual-level exploration for attention regression.   



\begin{figure}[h!]
\vspace{-4mm}
  \centering
  \includegraphics[width=0.45\textwidth]{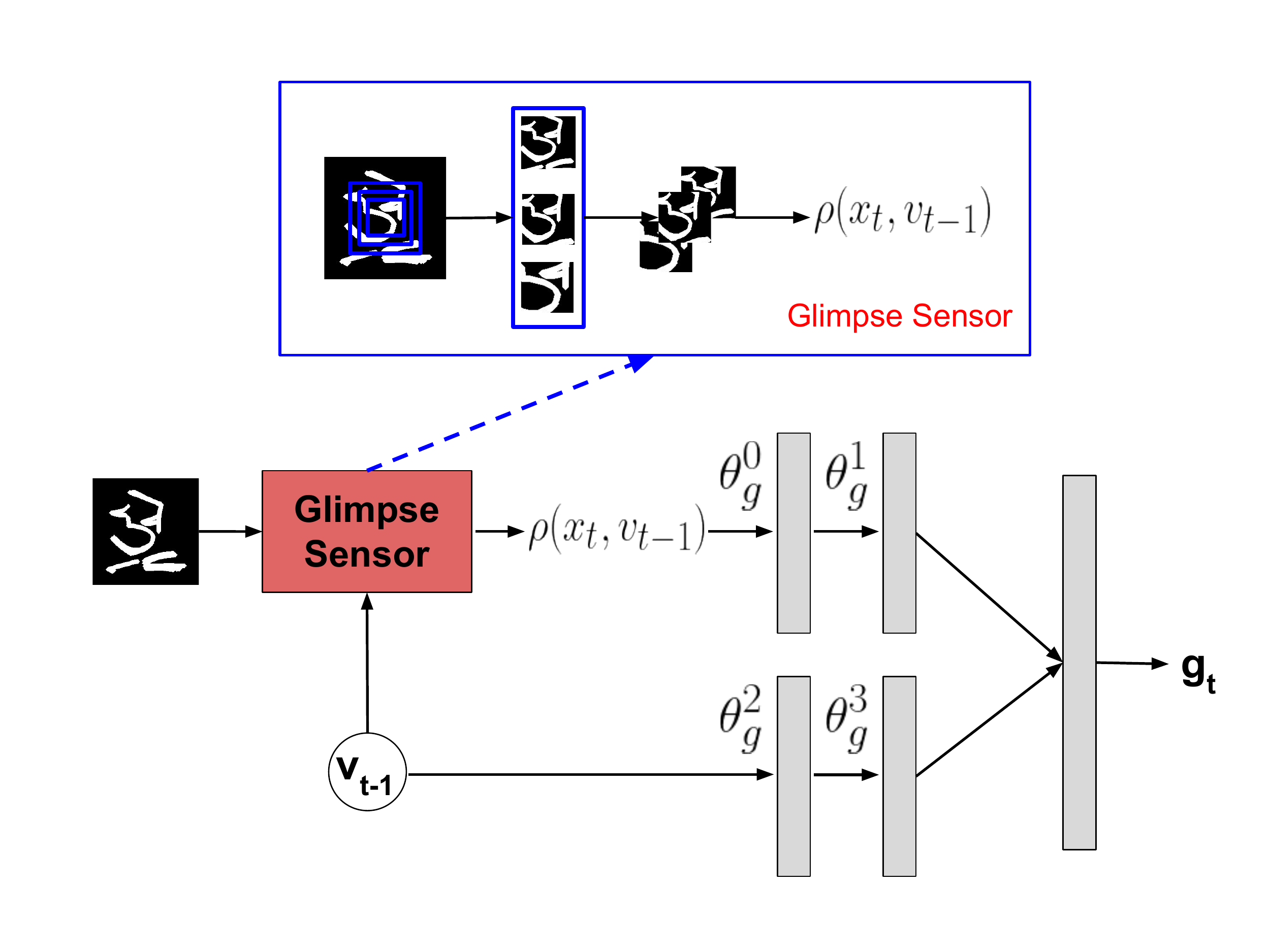}
\vspace{-4mm}
  \caption{The Glimpse Network} 
\vspace{-4mm}
\label{sensor}
\end{figure}

\vspace{1mm}\noindent{\bf Glimpse network.} As Figure \ref{sensor} shows, the glimpse network takes the original sketch ${\bf x}_i$  and a glimpse location vector ${\bf v}_{i,t-1}$ as input and  extracts the information of the retina-like representation
$\rho({\bf x}_i, {\bf v}_{i,t-1})$ around location ${\bf v}_{i,t-1}$ from sketch ${\bf x}_i$. It encodes the region around ${\bf v}_{i,t-1}$ at a high resolution but uses a progressively lower resolution for pixels farther from ${\bf v}_{i,t-1}$, resulting in a vector of lower dimensionality than the original sketch ${\bf x}_i$. The term \textit{glimpse} is to intuitively capture the combination of these high and low-resolution representations. The retina representation and glimpse location are then mapped into a hidden space using independent linear layers parameterized by $\theta_g = (\theta_g^0,\theta_g^1$; $\theta_g^2,\theta_g^3)$. With concatenation followed by rectified units, the glimpse network combines both representations to produce a final representation ${\bf g}_{i,t}$ to be used by the RNN network described next. 



\vspace{1mm}\noindent{\bf RNN network.} The RNN network takes the current time step's glimpse vector ${\bf g}_t$ and the previous hidden state vector ${\bf h}_{t-1}$ as input, and outputs the current time step hidden state vector ${\bf h}_{t}$. Here, the hidden state vector encodes the model’s knowledge of the environment and is instrumental in deciding how to act and where to deploy the glimpse sensor. The initial hidden vector is a random 256-dimensional embedding vector. The forward message of RNN is produced as follows: ${\bf h}_{i,t} = \text{RNN}({\bf h}_{i,t-1},{\bf g}_{i,t};\theta_h)$.


\vspace{1mm}\noindent{\bf Action Network.} The action network takes the hidden state ${\bf h}_t$ of the current time step as input and predicts the predicted final sketch representation: ${\bf a}_{i,t} = \text{ACTN}({\bf h}_{i,t};\theta_a)$. The predicted sketch representation is then used to calculate the distance from the corresponding image embedding vector as the supervised loss function: $\mathcal{L}^s_{i,t}=\|{\bf a}_{i,t}-{\bf e}_i\|^2$. It is worth to note that sketch representation is predicted using the current hidden state ${\bf h}_{i,t}$ from the RNN network, which has aggregated the previous hidden state and the current glimpse vector ${\bf g}_{i,t}$. Recall that the glimpse vector is computed from the currently attended region predicted by the located network (which will be described next) and the already completed partial sketch.  


\subsection{Dynamic Attention Regression via Dual-Level Exploration}
\vspace{1mm}\noindent{\bf Basic locator network.} The basic locator network takes the current hidden state ${\bf h}_{i,t}$ and maps it to a location that indicates an important region of the sketch that the model should attend to. To accommodate the on-the-fly retrieval of matching images where stokes may continue to be collected to make a partial sketch more complete, we propose to achieve dynamic attention regression through deep reinforcement learning. In particular, the basic locator network can be modeled as a policy network that learns a stochastic policy $\pi({\bf v}_{i,t}|{\bf h}_{i,t};\theta_v)$, where ${\bf v}_{i,t}\in \mathbb{R}^2$ contains the $(x,y)$ coordinates of the predicted location. The predicted location will then be used by the glimpse network to form the glimpse vector for the next time step along with the available partial sketch at that time step. Finally, the RNN will take the glimpse vector to transition (the environment) to the next hidden state ${\bf h}_{i,t+1}$. Since the action space corresponds the locations in the sketch image, which is continuous, we adopt a Gaussian policy:
\begin{align}
    {\bf v}_{i,t} \sim \mathcal{N}(\text{LOCN}({\bf h}_{i,t};\theta_v), \Sigma)
    \label{eq:gaussianpolicy}
\end{align}
where $\text{LOCN}({\bf h}_{i,t};\theta_v)$ is the mean location predicted based on the current hidden state and $\Sigma=\text{diag}(\sigma_x,\sigma_y)$. We will discuss how to set $(\sigma_x,\sigma_y)$ dynamically to achieve effective local exploration. 

\vspace{1mm}\noindent{\bf Dynamic ranking reward.} Since the ultimate goal of the sketch-based image retrieval is to rank the matching image higher than other irrelevant images, it is most appropriate to use a ranking based reward function. More intuitively, the reward should be large if the matching image is ranked high and small otherwise. 
Capturing this notion, we define the reward signal as follows:
\begin{equation}
    R_{i,t} = \mathbbm{1}(\text{score}_{i,t}\geq \eta_t)
    \label{eq: reward}
\end{equation}
where $\mathbbm{1} (\text{condition})$ is an indicator function taking value 1 if condition is true and 0 otherwise and $\eta_t$ is a dynamically adjusted threshold.
The score of the matching image $i$ is defined as the reciprocal of its ranking among all images:
\begin{equation}
    \text{score}_{i,t} = \frac{1}{\sum_{j=1}^N\mathbbm{1}(\|{\bf e}_j-{\bf a}_{i,t}\|^2_2\leq \|{\bf e}_i-{\bf a}_{i,t}\|^2_2)}
\label{rank}
\end{equation}
where ${\bf e}_i$ and ${\bf e}_j$  are the outputs of the image embedding network for images $i$ and $j$, respectively,  ${\bf a}_{i,t}$ is the predicted final sketch representation by the action network at time step $t$, and $N$ is the total number of images available in the training set.
In the early stage of a training episode, it is desirable to keep the threshold $\eta_t$ less strict, which allows the model to pick up the relatively weak reward signal. As the training progresses and the model gets better, the threshold is increased to encourage the model to seek for a higher reward.
To capture this notion, we define an adaptive threshold that depends on an training steps as follows:
\begin{equation}
    \eta_{t} = \frac{1}{1+\exp(-\alpha \times t+\beta)}
\label{eq:threshold}
\end{equation}
where $\alpha$ and $\beta$ are hyperparameters.

\vspace{1mm}\noindent{\bf Policy gradient.} Based on the reward signal defined above, the return at time-step $t$ is computed as the total discounted reward from $t$:
\begin{align}
G_{i,t}= \sum_{\hat{t}=1}^{T-\hat{t}} \gamma^{\hat{t}-1} R_{i,t+\hat{t}}   
\end{align}
where $0<\gamma\le 1$ is a discount factor that assigns a higher reward at an earlier stage, which has the effect to encourage the discovery of the matching image with a less complete partial sketch. Finally, we define the objective function for the basic locator network as the expected return (which is the state value) at time step $t$:
\begin{align}
    \mathcal{L}_{i,t}(\theta_v)=\mathbb{E}_{p({\bf h}_{i,t:T};\theta_v)}\left[\sum_{\hat{t}=0}^{T-t} \gamma^{\hat{t}} R_{i,t+\hat{t}}\right]
    \label{eq:reinforceloss}
\end{align}
where the hidden state distribution depends on the policy $\pi({\bf v}_{i,t}|{\bf h}_{i,t};\theta_v)$. Since the dynamics among the hidden environment states ${\bf h}_{i,t}$ is unknown, the expectation on the r.h.s. of \eqref{eq:reinforceloss} cannot be analytically computed. Thus, we perform Monte Carlo sampling to approximate the policy gradient:
\begin{align}
\nonumber \nabla_{\theta_v} \mathcal{L}_{i} (\theta_v) &= \sum_{t=1}^T \mathbb{E}_{p({\bf h}_{i,t:T};\theta_v)}\left[\nabla_{\theta_v} \log \pi({\bf v}_{i,t}|{\bf h}_{i,t}) G_{i,t}\right]\\
& \approx \frac{1}{M}\sum_{m=1}^M \sum_{t=1}^T \nabla_{\theta_v} \log \pi({\bf v}^{(m)}_{i,t}|{\bf h}^{(m)}_{i,t}) G^{(m)}_{i,t}
\label{eq:grad_reinforceloss}
\end{align}
where ${\bf v}^{(m)}_{i,t}, G^{(m)}_{i,t}$ are samples collected during episode $m$ for partial sketch $i$ at time step $t$ where  $m\in [M]$.

\vspace{1mm}\noindent{\bf Dual-level exploration.}
The partial sketch coupled with a potentially large image repository could make the reward signal very sparse and delayed. As a result, effective exploration is critical to ensure fast and better convergence for training the locator network. The recently developed Bootstrapped DQN model has been demonstrated to be a promising exploration strategy in discrete action space scenarios~\cite{Osband2016}. In particular, Bootstrapped DQN approximates the distribution over $Q$ values via a bootstrapping strategy to facilitate the deep and effective exploration. However, direct applying bootstrapped DQN to our SBIR framework faces two key challenges. First, it requires training two networks, including the $Q$ network for value approximation and a policy network to update the policy. Second, our action space is continuous, choosing the best action also requires solving an optimization problem. As a result, these challenges increase both the computational time and the difficulty of the training process, making it less suitable for the on-the-fly image retrieval.

 To deal with the key limitations presented above, we propose a dual level exploration technique that combines lightweight bootstrapped policy gradient (BPG) training for global deep exploration and adaptive local exploration through dynamically adjusting the variance of the Gaussian policy and the ranking reward. In particular, the BPG leverages a bootstrapped locator network with the $K$ heads to predict the glimpse location.
In each training episode, it randomly chooses one from $K$ candidate heads to predict the next location for the entire episode. To achieve the bootstrapping behavior, the experience $\{{\bf h}_{i,t}, {\bf v}_{i,t}, R_{i,t+1}, {\bf h}_{i,t+1}\}$ is recorded along with a mask $\boldsymbol{\omega}_{i,t} \in \{0,1\}^K$ that is sampled from a mask distribution $\Omega$. This experience is used to update the $k$-th locator network only when $\omega_{i,t}^k=1$. As a result, different heads will be trained using sufficiently different experiences to ensure the bootstrapping property. The proposed BPG offers three key advantages. First, by maintaining $K$ heads for the locator network and letting one head to execute one entire episode, it effectively achieves multi-step (or deep) exploration suitable for partial sketch matching with very sparse and delayed reward signals. Second, each head is trained using the bootstrapped experiences instead of from the same pool of experiences to effectively reduce the variance and improve the prediction accuracy. Last, by directly performing policy gradient, it avoids solving two optimization problems as in deep Q-learning to improve the efficiency when dealing with a continuous action space.

In addition to the global deep exploration using the $K$ heads, we equip each head with a local exploration capacity to ensure fast and better convergence of model training. In particular, the covariance matrix $\Sigma=\text{diag}(\sigma_x,\sigma_y)$ of the Gaussian policy in \eqref{eq:gaussianpolicy} plays an important to locate important regions in a sketch for the model to attend to. Within each head, if both $\sigma_x$ and $\sigma_y$ are set to very small, the head may be concentrated on a very small neighborhood and only performs exploitation without exploring other nearby regions. To perform effective local exploration, we propose to assign relatively large variances along with each coordinate in the early steps in the training and gradually shrink them along with the training process. Meanwhile, as shown in \eqref{eq:threshold}, the reward signal is also dynamically adjusted to allow the head to pick up a relatively weak signal in the early phase of training and then focus on higher rewards as the head becomes better trained. Our experimental results clearly justify the effectiveness of the dual-level exploration strategy.

\subsection{The Training Process}

For training, we use both supervised loss coupled with bootstrapped policy gradient loss. Given the input sketch ${\bf x}_i$ and glimpse location vector ${\bf v}_{i, t-1}$ from time step $t-1$, we use the glimpse network to determine the  glimpse representation ${\bf g}_{i, t}$. Next, in the RNN network, we pass ${\bf g}_{i, t}$ along with the previous hidden state representation ${\bf h}_{t-1}$ to produce the current hidden state vector ${\bf h}_{t}$. The action network then takes ${\bf h}_{t}$ and produces the corresponding sketch representation ${\bf a}_{i, t}$. Using this representation, we compute the distance from the corresponding image embedding vector as the supervised loss.  

On the other hand, given ${\bf h}_{t}$, the locator network with $K$ heads produces the next location ${\bf v}_{i, t}$ to be attended. Assume that the $k$-th head is selected to execute the current episode. The experience $\{{\bf h}_{i,t}, {\bf v}_{i,t}, R_{i,t+1}, {\bf h}_{i,t+1}\}$ is recorded along with a mask $\boldsymbol{\omega}^k_{i,t}$ to ensure the bootstrapping behavior during model training. Again, the network finds the glimpse considering the new location and repeats the process until the episode complete. It should be noted that we randomly choose the one $k$ head from $K$ candidate heads to predict the next location and it is fixed for the entire episode. We repeat the process for $M$ episodes. Finally, we update the locator network using \eqref{eq:grad_reinforceloss}. Next, we use the gradient of supervised loss to update the action network, RNN network, and glimpse network. More detailed training process is demonstrated in Algorithm~\ref{alg:DQN}.

  \begin{algorithm}[htb] 
  {\small
  \caption{Training Process for DARP-SBIR Framework 
  }  
  \label{alg:DQN}  
  \begin{algorithmic}[1]  
    \Require  
      Training sketch set ${\bf X}$, initial hidden state vector ${\bf H}_0$, initial glimpse location ${\bf V}_0$, pre-trained image embedding ${\bf E}$, total episodes $M$, Time step per episode $T$ 
    \Ensure  Updated network parameterrs $\theta_g, \theta_h, \theta_v, \theta_a$ 
    \State initialize supervised loss $\mathcal{L}^s$
    \label{start}
     \For {$m=0; m\leq M; m\leftarrow m+1$} 
      \State Sample $k \thicksim [1, .., K]$ 
      \For {$i=0; i\leq N; i\leftarrow i+1$} 
      \State Sample ${\bf x}_i \thicksim {\bf X}$
      \State Initialize: ${\bf v}_{i, 0}^{(m)}$, ${\bf h}_{i, 0}^{(m)}$
        \For {$t=0; t\leq T; t\leftarrow t+1$}
        \State Obtain retina representation $\rho({\bf x}_i, {\bf v}_{i, t}^{(m)})$ 
        \State Obtain ${\bf g}_{i, t}^{(m)}$ from Glimpse Network ($\rho({\bf x}_i, {\bf v}_{i, t}^{(m)}); \theta_g$) 
        \State Obtain ${\bf h}_{i, t+1}^{(m)}$ from $RNN ({\bf g}_{i, t}^{(m)}, {\bf h}_{i, t}^{(m)} ;\theta_h)$ 
        \State Obtain ${\bf a}_{i}^{(m)}$ from  $ACTN ({\bf h}_{i, t+1}^{(m)}; \theta_a)$ 
         \State Compute supervised loss and add to $\mathcal{L}^s$ 
        \State Compute $\mathcal{R}_{i, t+1}^{(m)}$ using \eqref{eq: reward} 
       \State Compute ${\bf v}_{i, t}^{(m)}$ using locator network $\theta_v$ 
        \State Store experience $\{{\bf h}_{i, t}^{(m)}, {\bf v}_{i, t}^{(m)}, \mathcal{R}_{i, t+1}^{(m)}, {\bf h}_{i, t+1}^{(m)}\}$ 
        \EndFor
        \EndFor

        \EndFor
        \State Update ${\theta}_v$  using \eqref{eq:grad_reinforceloss} 
        \State Update $\theta_a$, $\theta_h$, and $\theta_g$ using the supervised loss $\mathcal{L}^s$   
        \label{end}
        \State Repeat step 1$\thicksim$20
    
  \end{algorithmic}  
  }
\end{algorithm} 

\section{Experiments}\label{sec:exp}
In this section, we present our experimental results on three real-world sketch datasets to demonstrate: (1) the state-of-the-art performance in partial sketch based image retrieval, (2) effectiveness of attention regression through deep reinforcement learning, and (3) how the novel dual-level exploration helps to achieve better and faster convergence of the locator network that boosts the entire image retrieval performance. 

\subsection{Datasets}
We include three commonly used sketch datasets in our evaluation. The first two datasets, including QMUL-Shoe-V2\footnote{http://sketchx.eecs.qmul.ac.uk/downloads/}\cite{Pang2019,Song2018,Muhammad2018} and
 QMUL-Chair-V2\footnote{http://sketchx.eecs.qmul.ac.uk/downloads/}\cite{Song2018}, are specifically
 designed for FG-SBIR. Both datasets contain coordinate stroke information, enabling us to render the rasterized sketch images at intervals for training our DARP-SBIR framework and evaluate its retrieval performance over different stages of a complete sketch drawing episode. In our pre-processing step, we split the whole sketch drawing process into 17 partial sketches, using openCV's image dilation function to thicken the strokes, and leveraging the flipping and rotating function to perform data augmentation. We further include another publicly available fine-grained sketch-image dataset, Sketchy\footnote{https://sketchy.eye.gatech.edu/}\cite{Sangkloy2016}. We select 5 categories from this dataset  to test our framework's retrieval performance. Table \ref{tab0} summarizes the key properties of all three datasets.

 
\begin{table}[t!]
\caption{Description of Datasets}
\vspace{-2mm}
\begin{center}
\renewcommand\arraystretch{2}
\begin{tabular}{|c|c|c|c|c|c|c|c|c|c|}
\hline
\textbf{Dataset}&\multicolumn{3}{c}{\textbf{Sketch}}&\multicolumn{3}{|c|}{\textbf{Image}} \\
\cline{2-7} 
\textbf{Description} & \textbf{\textit{Train}}& \textbf{\textit{Test}}& \textbf{\textit{Total}}& \textbf{\textit{Train}}& \textbf{\textit{Test}}& \textbf{\textit{Total}}\\
\hline
ChairV2&725&275&1000&300&100&400 \\
\hline
ShoeV2&6051&679&6730&1800&200&2000\\
\hline
Sketchy&2500&500&3000&415&85&500\\
\hline

\end{tabular}
\label{tab0}
\end{center}
\vspace{-4mm}
\end{table}

\subsection{Comparison Baselines}
We include four competitive comparison baselines, where the first model primarily relies on a triplet network for both images and sketches, two other  models leverage relatively simple reinforcement leaning models to support partial sketch matching, and the last model that adapts the bootstrapped DQN into our proposed framework. We present the details of these models below:
\begin{itemize}[noitemsep,topsep=0pt,leftmargin=*]
    \item {\em Triplet network: } The triplet network outputs embedding vectors for both images and sketches \cite{Yu2016,Song2017}. By projecting both images and  sketches onto the embedding space, they can be directly matched through commonly used similarity or distance metrics. 
    \item {\em Triplet + Vanilla RL:} To support partial sketch matching, this model combines the triplet network for image embedding with a simple policy network~\cite{Mnih2016} to generate the final sketch representation , which can then be matched with the image embedding for retrieval. 
    \item {\em Triplet + PPO:} In this model~\cite{Bhunia2020}, the simple policy network is replaced by an improved version of the PPO network~\cite{Schulman2017}. The PPO network is trained with the replay memory. The value network is trained by Monte-Carlo method and the policy network is trained by policy gradient.
    \item {\em Bootstrapped DQN:} To demonstrate the effectiveness of the proposed deep attention model with dual-level exploration, we implement a Bootstrapped DQN \cite{Osband2016} version of the proposed framework by keeping all other network fixed but replacing the locator network with a Bootstrapped DQN for location regression.  
\end{itemize}

\subsection{Experimental Settings}
We conduct a grid search to set the value of some related parameters in the proposed framework. In particular,  we choose glimpse patches with a size of $8\times 8$ from the original $64\times 64\times 1$ input sketch and the number of glimpses per sketch is 12. The dimension for hidden glimpse vector ${\bf g}_{i,t}$ is 128, the dimension for hidden state vector ${\bf h}_{i,t}$ is 256, the sketch embedding dimension ${\bf a}_{i,t}$ is 64, same as the dimension of pre-trained image embedding ${\bf e}_i$. For the dynamic threshold given in \eqref{eq:threshold}, $\alpha$ is set as $0.02$ and $\beta$ as $-2$ (other values in a similar range also work well). The number of candidate heads in the bootstrapped policy network is set as $K=6$. The number of partial sketches in one episode is typically $17$. We train the model for 1,000 epochs to get desired results. Early stop is possible if the model doesn't improve the validation accuracy for a long time. The learning rate is set to be $3\times 10^{-4}$ and the resolution scale parameter is set to $1$. All the experiments are conducted on a 4-core Intel i7 CPU and one V100 GPU card with CUDA 10.2 and PyTorch 1.6.0. Our implementation source code can be found using this link\footnote{\url{https://github.com/wdr123/DARP-SBIR}}.


\subsection{Evaluation Metrics}
To properly evaluate the FG-SBIR performance under the on-the-fly setting, we consider two categories of evaluation metrics by following existing works in this area. In particular, to evaluate the retrieval performance based on complete sketches, we consider the images that appear at the top of the retrieved list as those that matter more. Therefore, our first metric evaluates the percentage of complete sketches with their true matching images appearing in the top-$q$ list, which is referred to as $acc@q$. To assess the retrieval performance on partial sketches, the metric should encourage early retrieval~\cite{Bhunia2020}. Thus, we compute the mean area under the plot of $1/\text{rank}$ versus the completion degree of a sketch, referred to as area under inverse rank or short for AUIR. We further consider the ranking percentile: $(N-\sum_{j=1}^N\mathbbm{1}(||{\bf e}_j-{\bf a}_{i,t}||^2_2\leq ||{\bf e}_i-{\bf a}_{i,t}||^2_2))/N$, versus the completion degree of a sketch as another metric for partial sketch performance evaluation.



\begin{figure*}[t!]
\begin{subfigure}{0.2\textwidth}
  \centering
  \includegraphics[width=1.0\linewidth]{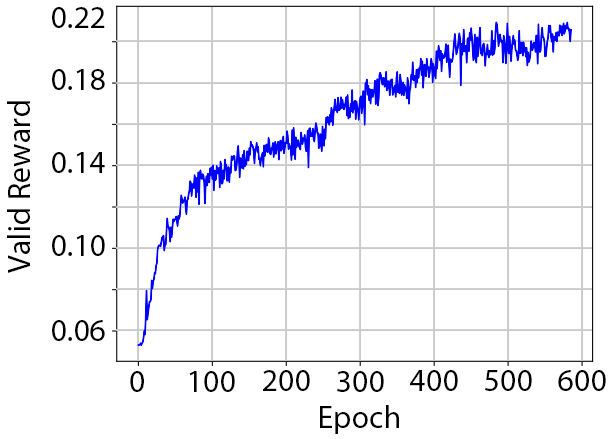}
   \vspace{-6mm}
  \caption{Triplet}
\end{subfigure}%
\begin{subfigure}{0.2\textwidth}
  \centering
  \includegraphics[width=1.0\linewidth]{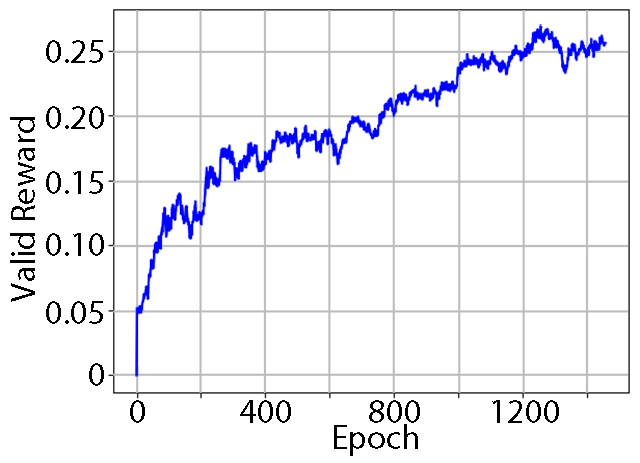}
\vspace{-6mm}
  \caption{Vanilla RL}
\end{subfigure}%
\begin{subfigure}{0.2\textwidth}
  \centering
  \includegraphics[width=1.0\linewidth]{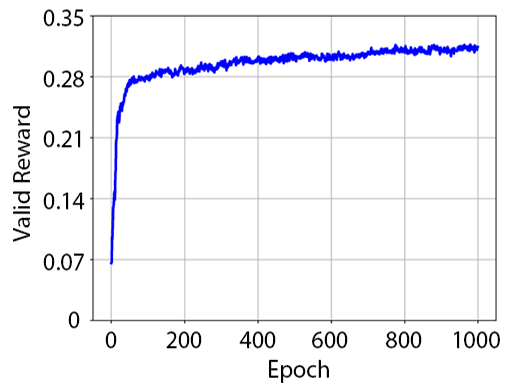}
\vspace{-6mm}
  \caption{PPO}
\end{subfigure}%
\begin{subfigure}{0.2\textwidth}
  \centering
  \includegraphics[width=1.0\linewidth]{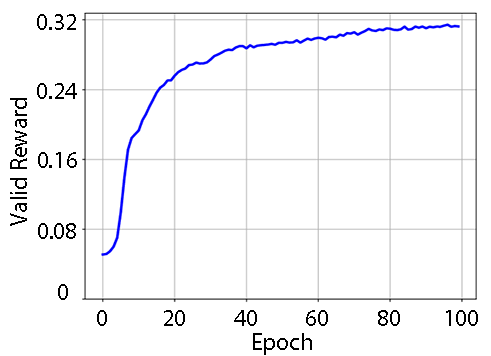}
  \vspace{-6mm}
  \caption{DQN}
\end{subfigure}%
\begin{subfigure}{0.2\textwidth}
  \centering
  \includegraphics[width=1.0\linewidth]{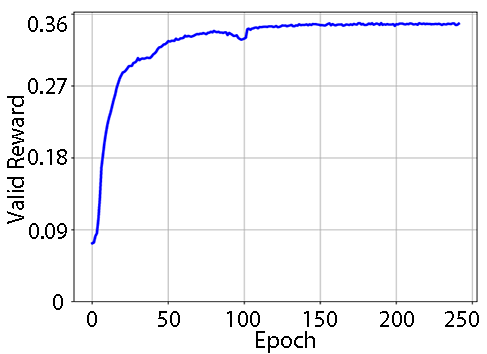}
\vspace{-6mm}
  \caption{DARP-SBIR}
\end{subfigure}%
\caption{Comparison Experiment }
\label{Comparison}
\end{figure*} 

\subsection{Comparison Results}

For the comparison within different baselines, we report the performance on
both partial and complete sketch based retrieval using AUIR and  $acc@5$ in Table~\ref{tab1}. 

\vspace{1mm} \noindent{\bf Complete sketch result.} First, for the complete sketch retrieval, the training curves of our model and other baselines on the ChairV2 dataset are shown in Figure \ref{Comparison}. The proposed model outperforms all the baselines with a clear margin. The second best model is the Triplet + PPO model that benefits from an advanced PPO network~\cite{Schulman2017}. However, no attention mechanism is provided, making it less robust to the noisy stokes. The Bootstrapped DQN based SBIR also performed worse than our framework. This could be due to the sub-optimal results obtained when optimizing over a continuous action space. For ChairV2 dataset, we leverage a pre-trained sketch embedding vector from \cite{Schulman2017} to initialize the RNN hidden state vector ${\bf h}_0$. For the other two datasets, we use a self-generated sketch embedding to initialize ${\bf h}_0$. Thus, all the baseline models are compared under the same condition in each dataset. 


\vspace{1mm} \noindent{\bf Partial sketch result.} For partial sketch retrieval, we use AUIR, which encourage early retrieval using less complete partial sketches. Benefiting from the deep reinforced attention regression mechanism, the proposed framework again outperforms all the baselines on all three datasets. The Triplet network, which is not designed for the on-the-fly retrieval, performs the worst in most cases.

\begin{figure}[htbp]
    \centering
    \begin{subfigure}[b]{0.25\textwidth}
    \centering
      \includegraphics[width=\textwidth]{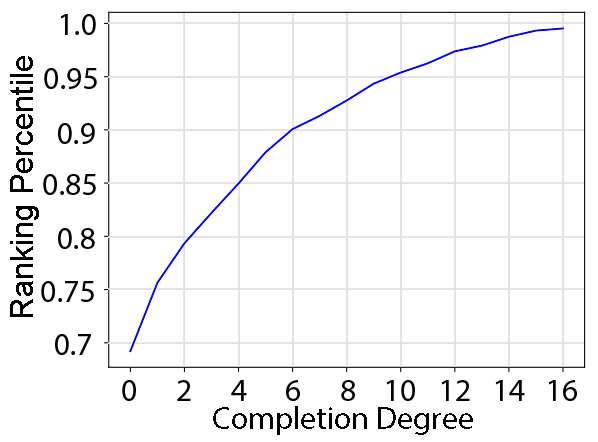}
      \vspace{-4mm}
      \caption{DARP-SBIR}
      
    \end{subfigure}%
    \begin{subfigure}[b]{0.25\textwidth}
      \centering
        \includegraphics[width=\textwidth]{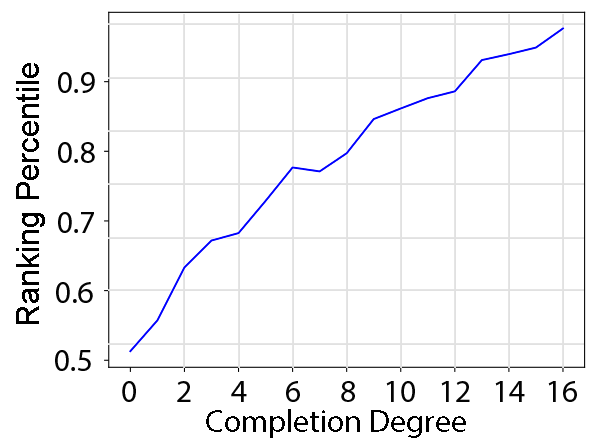}
        \vspace{-4mm}
        \caption{Triplet + PPO}
        
    \end{subfigure}%
    \caption{Ranking Percentile Vs. Complete Degree (ChairV2)}
    \label{rp_vs_partial}
    \vspace{-4mm}
 \end{figure}

Figure \ref{rp_vs_partial} further compares DARP-SBIR with the second best model, Triplet + PPO, on partial sketch retrieval through ranking percentile vs. complete degree of a sketch by using ChairV2 as an example. As can be seen, our model only needs six out of sixteen strokes to reach the top-10 ranking percentile while the Triplet + PPO needs more than 12 strokes to reach the same ranking. Performance on the other two datasets shows similar trends, which are omitted due to lack of space. 



\begin{table}[htbp]
\caption{Performance Comparison}
\vspace{-2mm}
\begin{center}
\renewcommand\arraystretch{2}
\resizebox{\hsize}{!}{
\begin{tabular}{|c|c|c|c|c|c|c|}
\hline
\textbf{Comparison}&\multicolumn{2}{|c|}{\textbf{Chair V2}}&\multicolumn{2}{|c|}{\textbf{Shoe V2}}&\multicolumn{2}{|c|}{\textbf{Sketchy}}\\
\cline{2-7}
\textbf{Model} & \textbf{\textit{AUIR}}& \textbf{\textit{acc@5}}& \textbf{\textit{AUIR}}&  \textbf{\textit{acc@5}} & \textbf{\textit{AUIR}}&  \textbf{\textit{acc@5}}\\

\hline
Triplet network&21.05&64.47&12.05&52.34&25.00&69.50\\
\hline
Triplet + Vanilla RL&25.34 &69.67 &13.60&54.80&24.30&68.70\\
\hline
Triplet + PPO&33.35  & 75.44 &15.50&57.10&27.88&71.24\\
\hline
Bootstrapped DQN&30.05  & 71.45 &13.50&54.50&24.50&69.00\\
\hline
\textbf{DARP-SBIR}& \textbf{35.65}&\textbf{79.12}&\textbf{18.12}&\textbf{60.02}&\textbf{32.32}&\textbf{73.82} \\
\hline

\end{tabular}}
\label{tab1}
\end{center}
\vspace{-2mm}
\end{table}\par

\begin{figure*}[h!]
\centering
\begin{subfigure}{0.24\textwidth}
  \centering
  \includegraphics[width=1.0\linewidth]{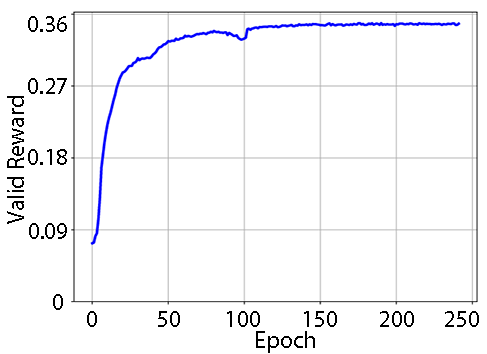}
  \vspace{-6mm}
  \caption{With Attention}
\end{subfigure}%
\begin{subfigure}{0.24\textwidth}
  \centering
  \includegraphics[width=1.0\linewidth]{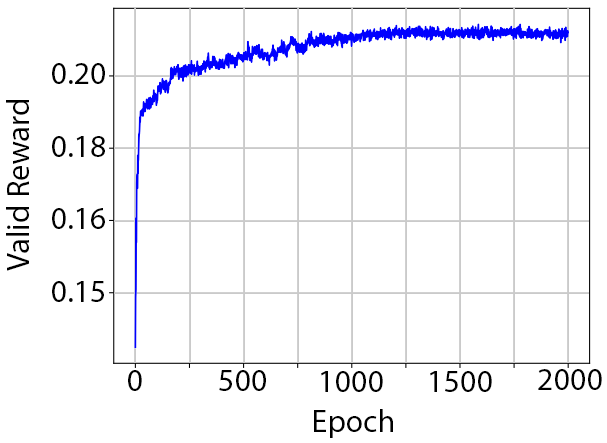}
\vspace{-6mm}
  \caption{W/O Attention}
\end{subfigure}%
\begin{subfigure}{0.24\textwidth}
  \centering
  \includegraphics[width=1.0\linewidth]{Ablation/ChairExpo/valid-reward_best.png}
\vspace{-6mm}
  \caption{With Exploration}
  \label{vreward65}
\end{subfigure}%
\begin{subfigure}{0.24\textwidth}
  \centering
  \includegraphics[width=1.0\linewidth]{Ablation/ChairExpo/valid_reward_hope5_new.png}
\vspace{-6mm}
  \caption{W/O Exploration}
  \label{vreward6}
\end{subfigure}%
\caption{Effectiveness of Attention Regression and Dual-Level Exploration}
\label{Ablation:Attention & Exploration}
\vspace{-2mm}
\end{figure*} 

 \begin{figure*}[h!]
    \begin{subfigure}{0.33\textwidth}
    \centering
    \includegraphics[width=\textwidth]{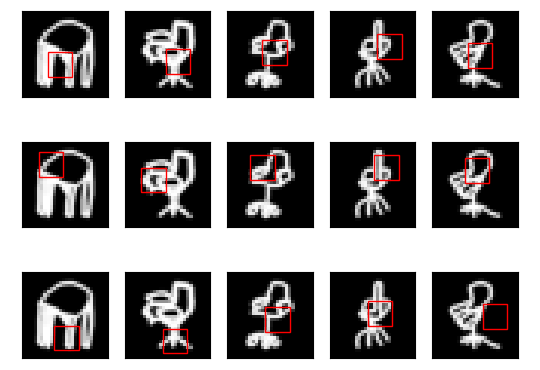}
    \caption{ChairV2}
    \end{subfigure}
    \begin{subfigure}{0.33\textwidth}
    \centering
    \includegraphics[width=\textwidth]{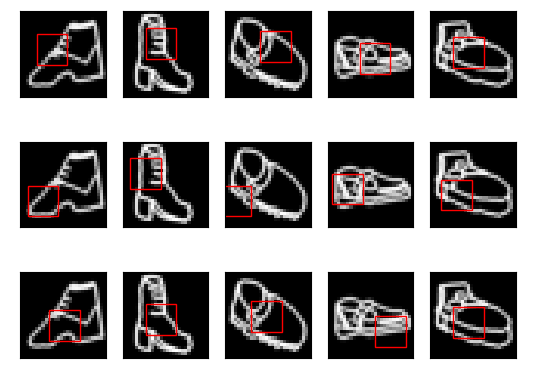}
    \caption{ShoeV2}
    \end{subfigure}
    \begin{subfigure}{0.33\textwidth}
    \centering
    \includegraphics[width=\textwidth]{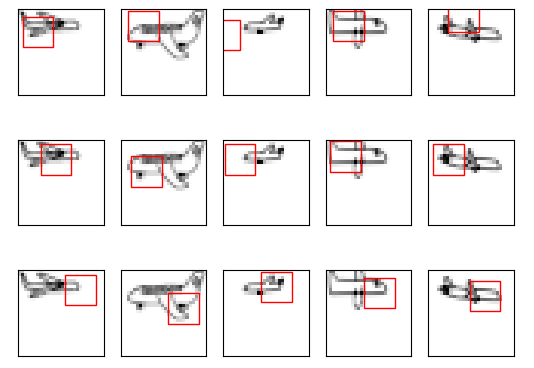}
    \caption{Sketchy}
    \end{subfigure}
    \caption{Glimpse Results\label{fig:glimpse}}
    \vspace{-2mm}
\end{figure*}


    
      
        

\subsection{Ablation Study}
We conduct an ablation study to evaluate the effectiveness of attention regression and the dual-level exploration used by the bootstrapped policy network.  

First, we compare the validation rewards of our model using two variants, with attention regression and without attention regression by using ChairV2 as an example. It is worth to note that without attention regression, both the glimpse network and the locator network will be removed from the framework. The RNN network is used to capture the temporal dynamics in the on-the-fly setting and the action network is to predict the final sketch representation. Figure \ref{Ablation:Attention & Exploration} (a-b) shows that the attention mechanism plays an important role in the overall retrieval performance by helping the framework identify the most important regions to attend to while ignoring the noisy stokes. Comparing to the model without attention selection and only using supervised loss, the model with attention mechanism gains around 75\% performance boost. Also, the model with attention mechanism could reach their maximum performance in relatively shorter time.





We further compare our model's performance with and without the dual-level exploration. Figure \ref{Ablation:Attention & Exploration} (c-d) shows that exploration is very important in performance boosting and reducing the training time. One thing to note, if a model only uses attention region selection while not applying sufficient exploration, then its performance is similar to not using attention region selection at all. This clearly demonstrate importance of exploration in partial sketch based image retrieval that involves  very sparse and delayed reward signals.

Finally, Figure \ref{fig:glimpse} shows the corresponding glimpse results for the three datasets to highlight the importance of local exploration (in combination with the global deep exploration). As shown, after proper training, our model is able to focus on the important regions of the sketches and therefore, ignoring the unnecessary strokes drawn. The attended regions usually cover a wide area in the sketches, partly due to the effect of dynamically adjusting the covariance in the Gaussian policy and the reward signal.

\section{Conclusions}
In this work, we propose a novel on-the-fly FG-SBIR framework that takes attention regions directly from an original sketch to achieve high sketch-based image retrieval performance. To deal with partial sketches and provide correct predictions as early as possible, it is trained with a bootstrapped policy network that performs novel dual-level exploration. To deal with the sparse and delayed reward signals in partial sketch based retrieval, we combine bootstrapping with a $K$-head locator network with effective local exploration strategies, including using an adaptive Gaussian covariance and dynamically adjusting the reward threshold. As a result, our framework can enforce an attention mechanism on the original sketch directly, reaching a high accuracy while not demanding a high computation cost. Experiments conducted on three real-world sketch datasets clearly demonstrate the state-of-the-art retrieval performance on both complete and partial sketch settings. 

\section*{Acknowledgement}
This research was supported in part by an NSF IIS award IIS-1814450 and an ONR award N00014-18-1-2875. The views and conclusions
contained in this paper are those of the authors and should not
be interpreted as representing any funding agency.

\bibliographystyle{IEEEtran}
\bibliography{reference}

\end{document}